\documentclass[sigconf,anonymous=False]{acmart}

\usepackage{algorithmic}
\usepackage{graphicx}
\usepackage{textcomp}
\usepackage{xcolor}
\usepackage{makecell}

\usepackage{multirow}
\usepackage{amsmath}
\mathchardef\mhyphen="2D 

\usepackage{siunitx}
\usepackage{pifont}
\usepackage{bm}
\usepackage{xcolor}  
\usepackage{soul}

\usepackage[font=small,skip=0pt]{caption}

\usepackage[ruled,%
            linesnumbered]{algorithm2e}
\SetKwInput{KwInput}{Input}

\AtBeginDocument{%
  }

\usepackage{fancyhdr}   

\fancypagestyle{firstpage}{     
  \fancyhf{}
  \chead{\textcolor{gray}{This article has been accepted for publication in the proceedings of the UbiComp/ISWC'25 Adjunct: Adjunct Proceedings of the 2025 ACM International Joint Conference on Pervasive and Ubiquitous Computing \& the 2025 ACM International Symposium on Wearable Computing}}
  \fancyfoot[C]{\small{\textcolor{gray}{~\copyright~ 2025 ACM.  Personal use of this material is permitted.  Permission from ACM must be obtained for all other uses, in any current or future media, including reprinting/republishing this material for advertising or promotional purposes, creating new collective works, for resale or redistribution to servers or lists, or reuse of any copyrighted component of this work in other works.}}}

}

\begin{document}

\title{TinierHAR: Towards Ultra-Lightweight Deep Learning Models for Efficient Human Activity Recognition on Edge Devices}

\author{Sizhen Bian}
\orcid{0000-0001-6760-5539}
\affiliation{%
  \institution{DFKI}
  \city{Kaiserslautern}
  \country{Germany}
}
\email{sizhen.bian@dfki.de}

\author{Mengxi Liu}
\affiliation{%
  \institution{DFKI}
  \city{Kaiserslautern}
  \country{Germany}
}
\email{mengxi.liu@dfki.de}

\author{Vitor Fortes Rey}
\affiliation{%
  \institution{DFKI}
  \city{Kaiserslautern}
  \country{Germany}
}
\email{vitor.fortes_rey@dfki.de}

\author{Daniel Geissler}
\affiliation{%
  \institution{DFKI}
  \city{Kaiserslautern}
  \country{Germany}
}
\email{daniel.geissler@dfki.de}

\author{Paul Lukowicz}
\affiliation{%
  \institution{DFKI}
  \city{Kaiserslautern}
  \country{Germany}
}
\email{paul.lukowicz@dfki.de}

\renewcommand{\shortauthors}{Sizhen Bian et al.}

\begin{abstract}

Human Activity Recognition (HAR) on resource-constrained wearable devices demands inference models that harmonize accuracy with computational efficiency. This paper introduces TinierHAR, an ultra-lightweight deep learning architecture that synergizes residual depthwise separable convolutions, gated recurrent units (GRUs), and temporal aggregation to achieve SOTA efficiency without compromising performance. Evaluated across 14 public HAR datasets, TinierHAR reduces Parameters by 2.7× (vs. TinyHAR) and 43.3× (vs. DeepConvLSTM), and MACs by 6.4× and 58.6×, respectively, while maintaining the averaged F1-scores. Beyond quantitative gains, this work provides the first systematic ablation study dissecting the contributions of spatial-temporal components across proposed TinierHAR, prior SOTA TinyHAR, and the classical DeepConvLSTM, offering actionable insights for designing efficient HAR systems. We finally discussed the findings and suggested principled design guidelines for future efficient HAR. To catalyze edge-HAR research, we open-source all materials in this work for future benchmarking\footnote{https://github.com/zhaxidele/TinierHAR}.

\end{abstract}




\maketitle

\section{Introduction}

\thispagestyle{firstpage} 

Human Activity Recognition (HAR) has emerged as a cornerstone technology for enabling context-aware applications in fields like healthcare \cite{serpush2022wearable, luder2024cyclowatt, amft2020wearables}, fitness tracking \cite{phukan2022convolutional,bian2019passive}, smart environments \cite{igwe2022human}, and assistive robotics \cite{fiorini2021daily, liu2022non}. By inferring human activities from sensor data captured via wearable or embedded devices, HAR systems empower real-time decision-making \cite{xu2021human,bonazzi2024retina}, personalized interventions \cite{mhalla2024domain,kang2024device}, and enhanced user experiences \cite{zhang2022deep,bian2021systematic}. 
However, deploying HAR models on resource-constrained edge devices introduces a critical challenge: achieving high recognition accuracy while adhering to stringent computational, memory, and energy constraints \cite{muhoza2023power}. 
Traditional deep learning architectures, though accurate, often prove infeasible for on-device inference due to their excessive parameter counts and computational demands\cite{ronald2021isplinception}. Conversely, overly simplified models risk sacrificing performance, limiting their practical utility in real-world scenarios\cite{bian2021capacitive}. 
Recent advancements in lightweight HAR neural architectures, such as TinyHAR\cite{zhou2022tinyhar} and MLPHAR\cite{zhou2024mlp}, have sought to address the trade-off of computational demands and inference performance by optimizing model efficiency. Yet, we noticed that there is still a large volume to further push the envelope of resource usage while synergistically maintaining or even enhancing modeling.

This work introduces TinierHAR, an ultra-lightweight architecture designed to achieve state-of-the-art efficiency without compromising accuracy for HAR. Building on the synergized residual depthwise separable convolutions (to capture hierarchical sensor signal patterns with minimal parameters), gated recurrent units (to efficiently encode sequential dependencies), and temporal aggregation (to reduce redundancy in temporal feature maps), TinierHAR optimizes spatial-temporal feature learning while significantly minimizing computational overhead. 
We evaluate TinierHAR on 14 publicly available HAR datasets, representing the largest scale evaluation of its kind to date, encompassing a wide range of sensor modalities, activity categories, and application domains. Compared to TinyHAR and DeepConvLSTM, TinierHAR achieves an average reduction in model size by 2.7× and 43.3×, respectively, and reduces multiply-accumulate operations (MACs) by 6.4× and 58.6×, while matching their average F1-scores. Beyond these quantitative gains,  we also contribute: (1) a comprehensive ablation study to dissect the impact of each architectural component across the three models; (2) a systematic scaling analysis to explore the scaling laws in aspect of model size for HAR applications; and (3), an in-depth discussion about the findings based on the quantitative result and future directions. We hope that such takeaways could offer actionable insights for designing future efficient HAR systems. 


\begin{figure*}[hbt]
\centering
\includegraphics[width=0.75\linewidth, height = 2.9cm]{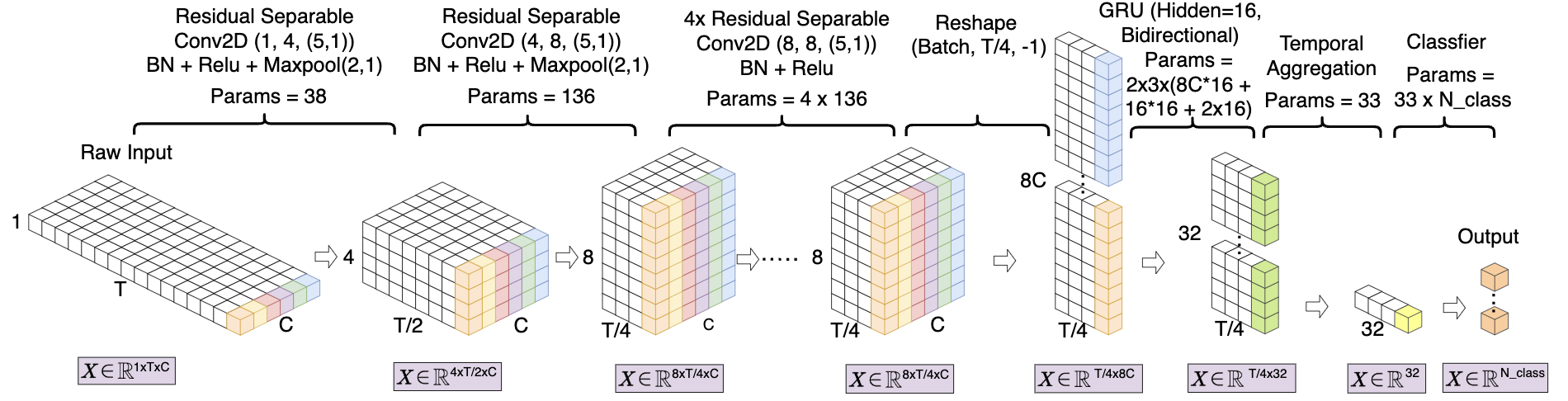}
\caption{The architecture of TinierHAR}
\label{Arch}
\end{figure*}

\section{Related Work}

Deploying HAR models on edge devices has spurred research into efficiency-oriented architectures. A common strategy involves replacing standard convolutions with depthwise separable convolutions\cite{zhu2021lightweight, bian2024device}.
Recent studies \cite{lalwani2024novel,batool2024ensemble} also suggest that traditional recurrent networks remain popular for temporal modeling in HAR due to their strong ability to capture sequential dependencies. However, their computational overhead has motivated alternatives such as temporal pooling \cite{song2018deep} and dilated convolutions \cite{hamad2021dilated, bian2025hybrid}. 
Work like \cite{kang2025human} proposed pruning GRU layers to reduce latency, but this often sacrifices temporal granularity critical for fine-grained activity recognition. 
Beyond architectural innovations, techniques like quantization \cite{yi2023human}, knowledge distillation \cite{deng2023lhar}, and neural architecture search (NAS) \cite{lim2023efficient} have been applied for efficient HAR. 
Recent work from KIT introduced TinyHAR \cite{zhou2022tinyhar}, a lightweight architecture that combines temporal convolutional blocks, self-attention mechanisms, LSTM layers, and temporal aggregation to optimize both spatial-temporal feature extraction and efficiency. TinyHAR achieves competitive accuracy while significantly reducing model size and computational costs.
Despite advancements in recent lightweight HAR research, a few critical domain knowledge remain unaddressed: (1) Insufficient Methodological Rigor: prior works lack granular analyses of architectural component efficacy, impeding evidence-based model design; (2) Limited Robustness: existing architectures exhibit weak generalizability explorations across heterogeneous sensor configurations and activity taxonomies. (3) Inadequate Scalability: considering diverse edge-device constraints, current approaches seldom systematically explore scalable architectures that balance accuracy and efficiency.

To address those limitations, we propose TinierHAR, 
pushing the frontier of edge HAR in both efficiency and performance. Our work advances the field through: (1) An ultra-lightweight deep learning architecture for HAR that combines residual depthwise separable convolutions with bidirectional GRU and temporal aggregation reaching the SOTA tradeoff of performance and efficiency among its kind; (2) The first systematic ablation study quantifying component-level contributions in different architectures across 14 datasets and the validation of architectural scalability. (3) The in-depth explanation of the findings of edge HAR based on the qualitative result and a few insights on future directions for principled design guidelines. 

\section{Architecture Design}
TinierHAR is designed for edge HAR and integrates spatiotemporal feature learning, temporal sequence modeling, and adaptive aggregation into a cohesive framework. The architecture is modular, enabling systematic ablation studies and scalability across diverse sensor configurations (Figure \ref{Arch}). 
The model is composed of the following units:

\noindent\textbf{Hierarchical Spatial Feature Extraction with Residual Separable CNN}: The spatial processing backbone employs a cascade of Residual Depthwise Separable Convolution blocks to capture multi-scale motion patterns while minimizing computational costs. Each block replaces standard convolutions with a two-step operation: the depthwise convolution, which applies a temporal kernel independently to each input channel to extract local features; And the pointwise convolution, which merges cross-channel correlations via 1×1 convolutions. Each block includes a residual shortcut to mitigate gradient degradation: using 1×1 convolutions with batch normalization when channels of input and output are different, otherwise, an identity mapping. To further enhance efficiency, the first two convolutional blocks apply stride-2 max pooling, halving the temporal dimension and reducing a significant amount of computational costs.

\begin{figure*}[hbt]
\centering
\includegraphics[width=0.99\linewidth, height = 7.0cm]{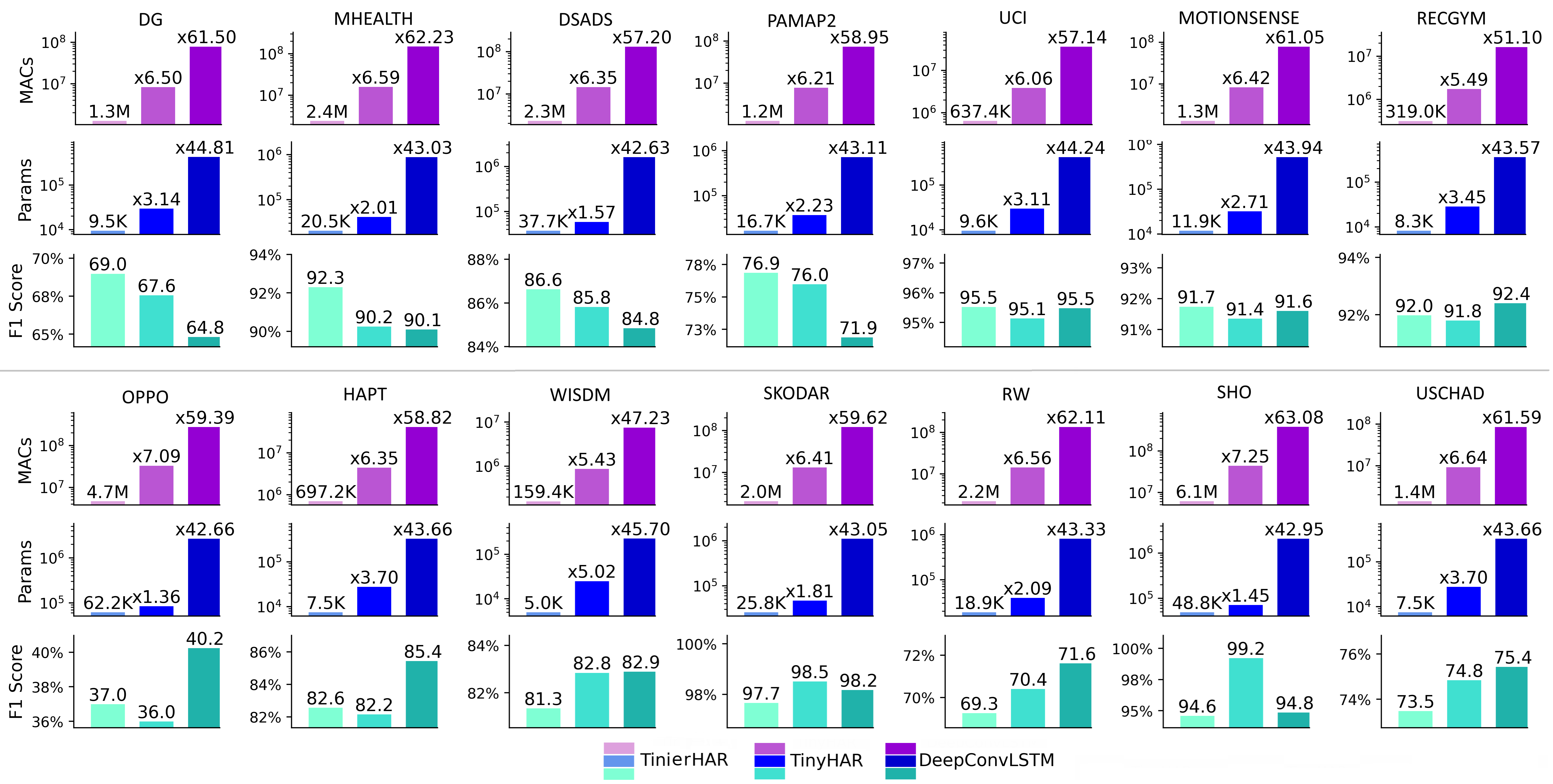}
\caption{Evaluation result of TinierHAR on 14 HAR datasets. Averages across datasets: 2.7x, 43.3x in parameters, 6.4x, 58.6x in MACs, 1.000x, 1.006x in macro F1-Score of TinyHAR and DeepConvLSTM compared with Tinierhar}
\label{Result}
\end{figure*}

\noindent\textbf{Lightweight Temporal Modeling with Bidirectional GRU}: The CNN output is flattened into a temporal sequence and processed by a bidirectional GRU to model temporal dependencies in human activities. Bidirectional GRU enhances temporal modeling by capturing both past and future context, improving recognition of complex motion sequences. Compared to LSTM, the GRU has faster inference time due to the absence of an additional cell state and fewer gating operations. The bidirectional mechanism doubles the temporal receptive field without significantly increasing memory costs, as weights are shared across both directions. 

\noindent\textbf{Attention-Based Temporal Aggregation}: Instead of using static pooling to summarize temporal features, TinierHAR employs an attention-based temporal aggregation mechanism that dynamically weights the most informative time steps: a learnable linear layer followed by the Softmax first computes importance scores for each time step and the context-weighted sum is then calculated as the final feature vector. This allows the model to focus on key activity segments while discarding redundant information. Such adaptive importance weighting enables the model to learns a set of attention scores for each time step, emphasizing frames that contribute most to activity classification while down-weighting irrelevant segments. 
By replacing traditional pooling methods with learned attention, TinierHAR achieves a more discriminative representation of activity sequences, improving performance without additional computational cost.

\begin{table}[]
\centering
\caption{14 Evaluated HAR Datasets}
\label{datasets}
\begin{tabular}{ p{1.50cm} p{2.1cm} p{0.55cm} p{0.50cm}  p{0.55cm} p{0.65cm} p{0.35cm} }
\hline
Datasets & Sensor & Subj- ects & Clas- ses & Chan- nels & Frequ- ency & SW$^{a}$\\
\hline
DG\cite{DG_Dataset} & 3xAcc & 10 & 9 & 9 & 64 & 4 \\
USCHAD\cite{USCHAD_Dataset} & 1xAcc/Gyro & 7 & 12 & 6 & 100  & 4\\
SKODAR\cite{SKODAR_Dataset} & 10xAcc & 1 & 10 & 30 & 33  & 4 \\
PAMAP2\cite{PAMAP2_Dataset} & 2xAcc/Gyro/Mag & 9 & 12 & 18 & 33  & 4 \\
DSADS\cite{DSADS_Dataset} & 5xAcc/Gyro/Mag & 8 & 19 & 45 & 25  & 4 \\
HAPT\cite{HAPT_Dataset} & 2xAcc/Gyro & 10 &  12 & 6 & 50  & 4\\
RW\cite{RW_Dataset} & 7xAcc & 15 & 8 & 21 & 50  & 4 \\
WISDM\cite{WISDM_Dataset} & 1xACC & 36 & 6 & 3 & 20  & 4 \\
OPPO\cite{OPPO_Dataset} & 7xAcc/Gyro/Mag, 2xMag, 2xQuanternion & 4 & 18 & 77 & 30  & 4 \\
RECGYM\cite{RECGYM_Dataset} & 1xAcc/Gyro, 1xCapative & 10 & 7 & 12 & 20 & 4  \\
MOTION- SENSE\cite{MOTIONSENSE_Dataset} & 1xAcc/Gyro & 24 & 12 & 6 & 50 & 4  \\
MHEALTH\cite{MHEALTH_Dataset} & 2xAcc/Gyro/Mag, 1xACC, 2xECG & 10 & 12 & 23 & 50 & 4 \\
SHO\cite{SHO_Dataset} & 5xAcc/Gyro/Mag, 5xLACC & 10 & 7 & 60 & 50  & 4 \\
UCI\cite{UCI_Dataset} & 1xAcc/Gyro/LACC & 30 & 6 & 9 & 50 & 4 \\

\hline
\end{tabular}
\begin{minipage}{\textwidth}
    \setlength{\columnsep}{1.8cm}
    $^{a}$ Sliding window of T seconds with T/2 overlap. For consistency, \newline we apply T=4s across all datasets for the evaluation.
\end{minipage}
\end{table}

\section{Experimental Results}

TinierHAR was evaluated across 14 publicly available HAR datasets (Table \ref{datasets}), comparing its computational efficiency and performance against TinyHAR and DeepConvLSTM. These two models were selected for the following reasons: (1) TinyHAR is currently the SOTA in efficient HAR. While MLP-HAR demonstrates slightly better efficiency, it is not an end-to-end model requiring FFT preprocessing. 
(2) DeepConvLSTM is one of the most widely recognized architectures in HAR, and its robustness and reliability have been extensively validated over the past decade. 

All datasets were segmented into 4-second windows with 2-second overlap for consistency. The training was on Nvidia RTX A6000. To ensure reproducibility and consistent results, the initial parameters were generated with five seeds (1-5), and the average was reported. During training, the epoch is set to 150 with early stopping (patience = 15), the learning rate is set to 0.001 with a degradation factor of 0.1 and patience of 7,  the loss criterion is CrossEntropy and the optimizer is AdamW. For most of the datasets, the leave-one-user-out strategy was applied to guarantee a robust result (a few exceptions like WISDM, MOTIONSENSE, and UCI, subjects are grouped first; and for SKODAR, since only one subject participated in the experiments, the leave-one-session-out strategy is applied). The evaluation metrics include MACs and Parameters, and Macro F1-score. The detailed results are visualized in Figure \ref{Result}.

In the aspect of computational efficiency, TinierHAR exhibits a remarkable reduction in computational cost compared to its predecessors. Regarding MACs, TinierHAR achieves an average decrease of 6.4× compared to TinyHAR and 58.6× compared to DeepConvLSTM, significantly improving inference efficiency. Similarly, in parameter efficiency, TinierHAR requires 2.7× fewer parameters than TinyHAR and 43.3× fewer than DeepConvLSTM, reinforcing its lightweight design. Across all datasets, TinierHAR consistently maintains its computational advantage, with the most substantial reductions observed in SHO, RW, MHEALTH, USCHAD, and DG, highlighting its effectiveness in handling complex activity patterns with significantly reduced resource requirements.


\begin{figure*}[hbt]
\centering
\includegraphics[width=0.89\linewidth, height = 5.5cm]{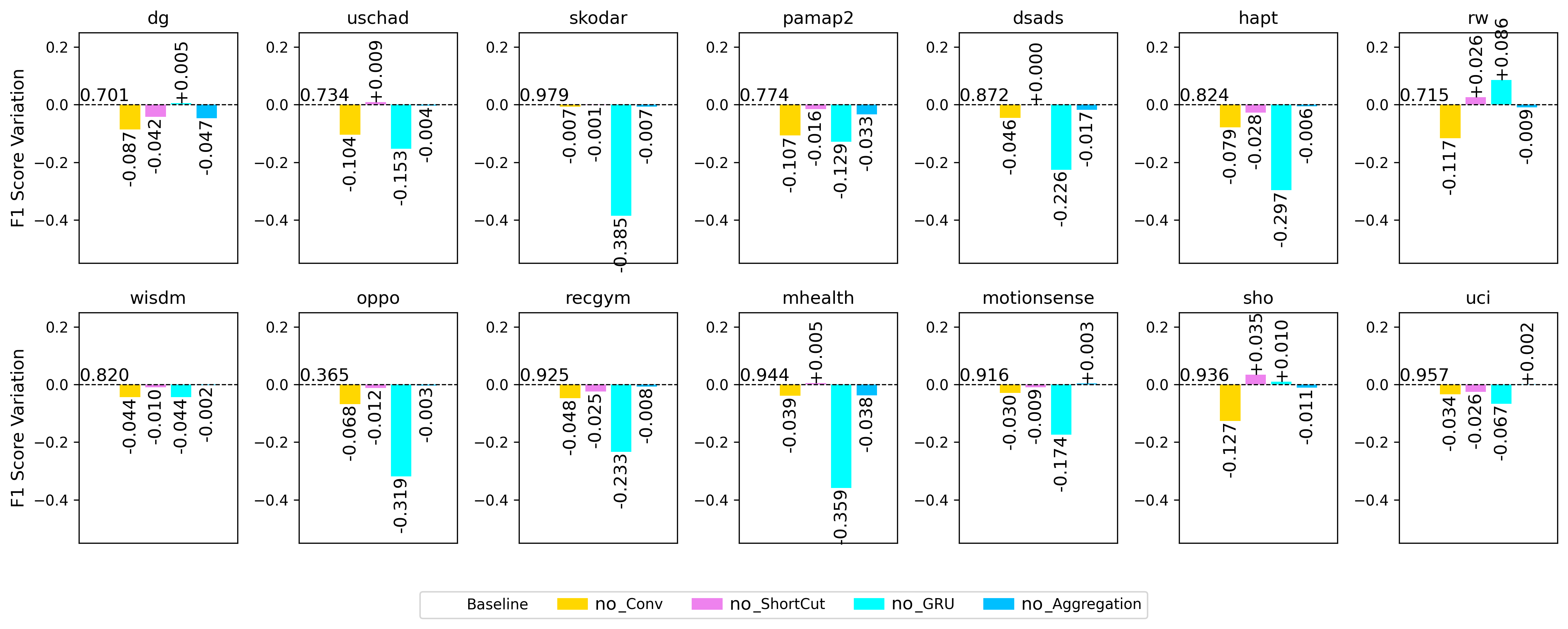}
\caption{F1-Score Variation of TinierHAR Ablation Study Across Datasets. Averaged variations are -9.19\%, -0.95\%, -19.63\%, -1.74\% when removing the convolutional block, shortcut block, GRU block, and aggregation block, respectively.}
\label{AS_TinierHAR}
\end{figure*}

\begin{figure*}[hbt]
\centering
\includegraphics[width=0.89\linewidth, height = 5.5cm]{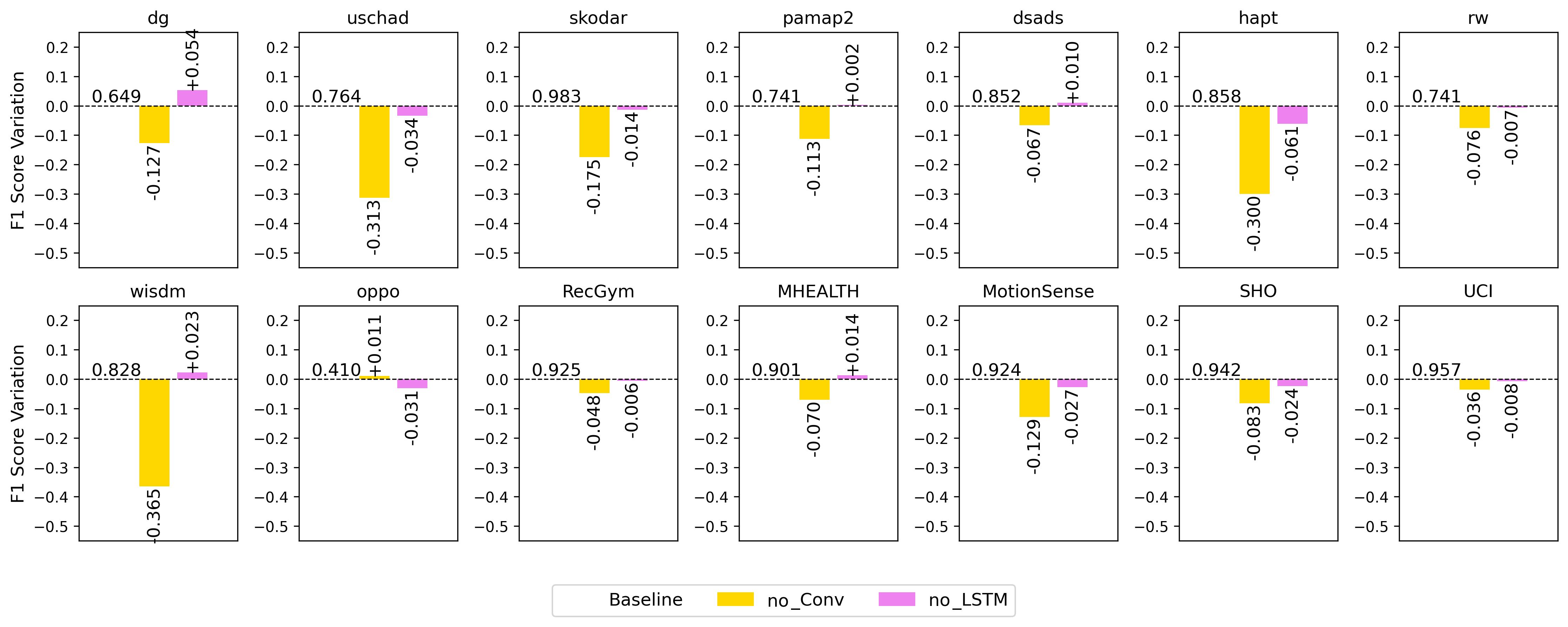}
\caption{F1-Score Variation of DeepConvLSTM Ablation Study Across Datasets. Averaged variations are -27.50\%,
-1.84\% when removing the convolutional block and LSTM block.}
\label{AS_DeepConvLSTM}
\end{figure*}

In the aspect of recognition performance, despite its drastically lower computational cost, TinierHAR maintains remarkable recognition accuracy across a diverse set of HAR datasets: outperforming in 6 out of 14 datasets(DG, MHEALTH, DSADS, PAMAP2, UCI, MOTIONSENSE). Meanwhile, DeepConvLSTM leads in 6 datasets (RECGYM, OPPO, HAPT, WISDM, RW, USCHAD), whereas TinyHAR only leads in SKODA and SHO. When assessing the relative F1-score variation across models, defined as:
\begin{math}(F1\mhyphen Score_{tinyhar/deepconvlstm}-F1\mhyphen Score_{tininerhar})/F1\mhyphen Score_{tinierhar}\end{math},
we observe that TinierHAR achieves an identical macro F1-score on average compared to TinyHAR  (1.000×). Furthermore, while DeepConvLSTM slightly outperforms TinierHAR on average (1.006×), this comes at a massive cost in computational complexity, reinforcing that Parameter and MAC reductions in TinierHAR do not compromise recognition accuracy.

\section{Discussion}


Building on the quantitative observations desribed above, this section outlines two further studies and key findings designed to further investigate the underlying principles and contributing factors of efficient human activity recognition (HAR). By systematically analyzing architectural design choices, computational trade-offs, and performance trends across various models and datasets, we aim to uncover critical insights that can guide the development of more effective and resource-efficient HAR systems.

\begin{figure*}[hbt]
\centering
\includegraphics[width=0.89\linewidth, height = 5.5cm]{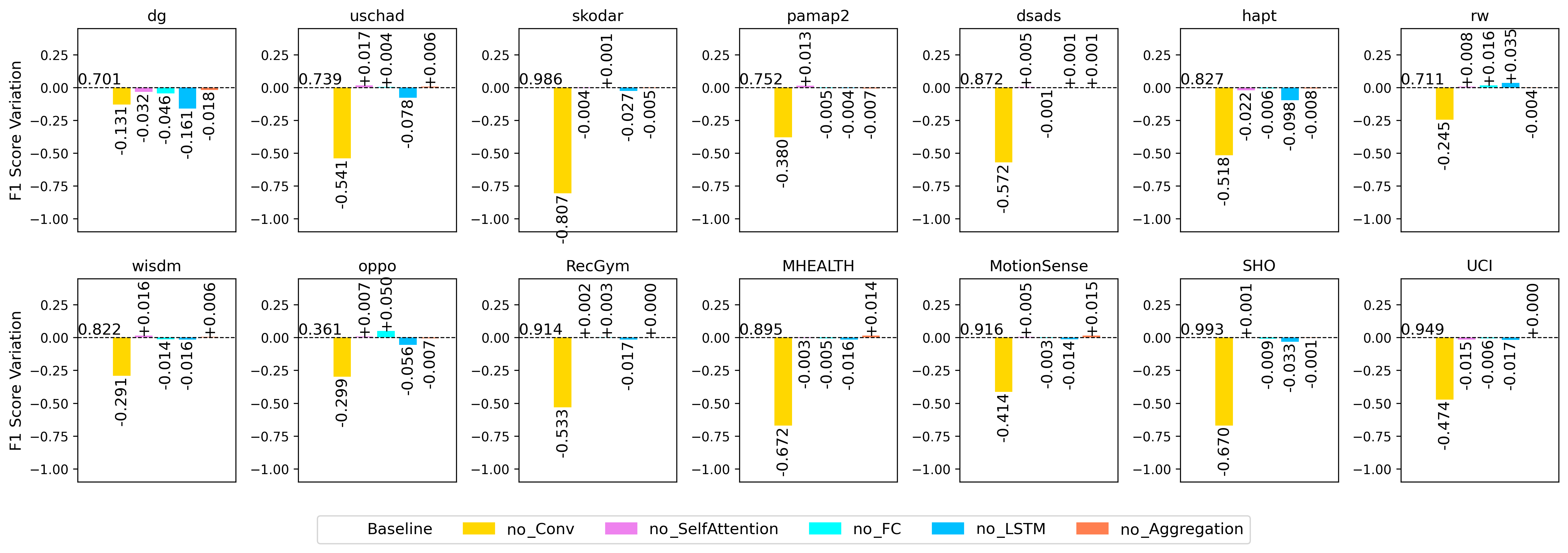}
\caption{F1-Score Variation of TinyHAR Ablation Study Across Datasets. Averaged variations are -55.06\%, +0.25\%, +0.40\%, -5.26\%, -0.02\% when removing the convolutional block, self-attention block, fully connected block, LSTM block, and aggregation block, respectively}
\label{AS_TinyHAR}
\vspace{-0.5em}
\end{figure*}

\begin{figure*}[hbt]
\centering
\includegraphics[width=0.9\linewidth, height = 4.0cm]{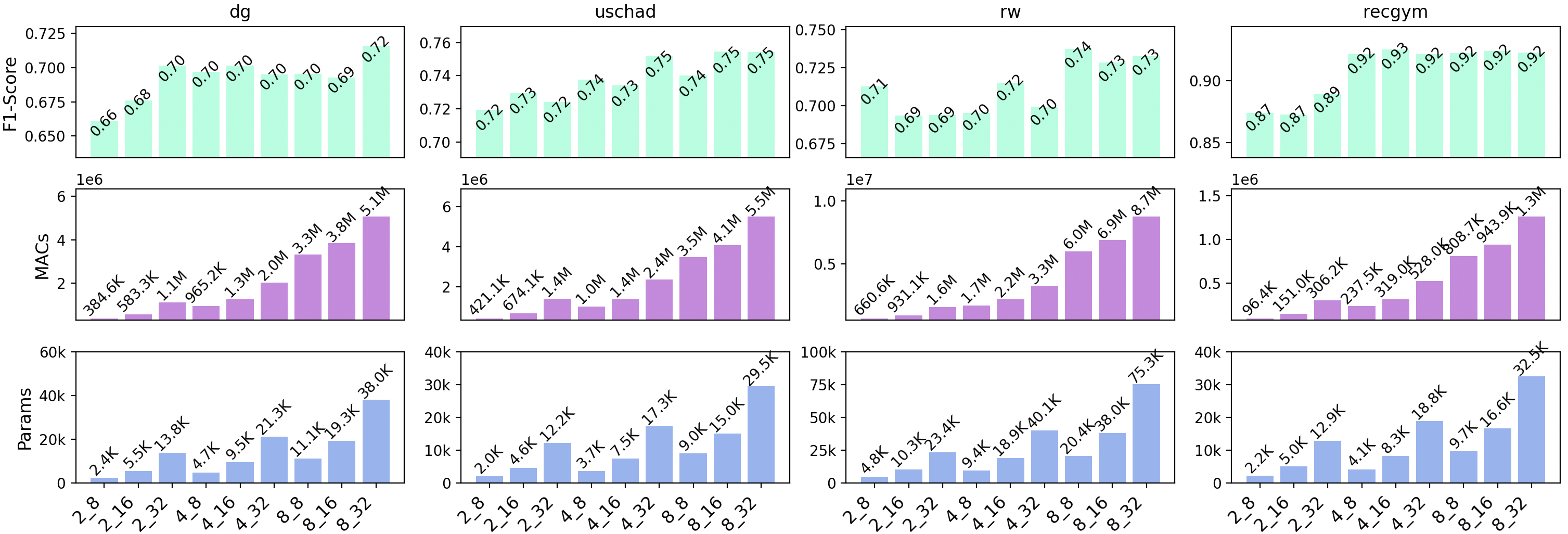}
\caption{Metrics Across Models Configurations and Datasets (M\_N of the labels in bottom x-axis means M residual separable convolutional blocks (besides the first two with pooling) with M filters and N hidden size for the GRU block).}
\label{scaling}
\vspace{-0.5em}
\end{figure*}

\subsection{Ablation Study}

Figure \ref{AS_TinierHAR} presents the F1-Score variations in the TinierHAR ablation study across the datasets, showing the F1-score baseline and the relative change when each component is removed. 
Removing the CNN results in an average F1-score drop of -9.19\%, indicating that this component significantly contributes to feature extraction. The impact is especially noticeable in datasets like SHO (-12.7\%) and RW (-11.7\%). 
The removal of shortcut connections leads to a relatively small average drop of -0.95\%, suggesting that while residual connections improve gradient flow, their absence does not drastically affect overall performance. While most datasets show a more or less drop when removing the shortcut, the SHO and RW datasets even show a slight improvement (+3.5\%, +2.6\%), indicating that shortcut paths may not always be necessary for specific datasets. 
Removing the GRU component has the most severe negative impact (-19.63\% on average), particularly in SKODAR (-38.5\%) and MHEALTH (-35.9\%), confirming that GRUs are essential for capturing temporal dependencies in sensor data. Interestingly, for the RW datasets, the GRU unit shows a negative effect; it will get the best performance of 80.1\% when removing the GRU unit. The reason might be located in the less complex temporal patterns of the signals in the RW dataset, and spatial relationships captured by CNNs are more important than temporal ones.
The removal of the aggregation block leads to a moderate average F1-score drop of -1.74\%, indicating that this module plays a role in refining final predictions, but its effect varies across datasets. 
In summary, GRU Block is the most critical, as its removal leads to the most severe performance degradation, proving its necessity for temporal feature learning. CNN Block plays a significant role in feature extraction, with notable F1-score drops when removed. Aggregation and Shortcut Blocks contribute to stability, but their impact varies depending on the dataset. Some datasets are more dependent on specific blocks than others, emphasizing the need for a dataset-specific architecture.

Figure \ref{AS_DeepConvLSTM} illustrates the F1-score variation from an ablation study of the DeepConvLSTM model. Across almost all the datasets, removing the convolutional block leads to a significant drop in F1-Score, with some datasets (WISDM, USCHAD, HAPT) experiencing a drop of more than 30\%. This indicates that CNN layers are crucial in capturing local spatial dependencies in sensor signals. In most cases, removing the LSTM block results in a minimal decrease or even a slight improvement in performance (DG, PAMAP2, WISDM, DSADS, MHEALTH). This suggests that temporal dependencies captured by LSTM are not always essential, possibly because the convolutional layers already extract meaningful sequential patterns. However, some datasets like USCHAD and HAPT show minor negative effects when removing LSTMs, indicating that LSTM blocks may still be helpful for specific datasets with complex temporal relationships. The results highlight that DeepConvLSTM’s strength comes primarily from its CNN layers, while the LSTM blocks may offer only marginal benefits for some datasets (datasets with strong spatial dependencies rely more on CNNs, while those with long-term temporal dependencies might benefit from LSTMs). 


Figure \ref{AS_TinyHAR} presents the F1-score variation across datasets when removing different components from the TinyHAR model. Across all datasets, eliminating the convolutional layers results in a massive drop in F1-Score, often by more than 50\% (e.g., SKODAR, HAPT, MHEALTH, SHO, RECGYM). This confirms that CNN layers are the primary feature extractors. Interestingly, eliminating the self-attention mechanism does not lead to a significant drop. In fact, up to 10 datasets out of the 14 show slight performance improvement, indicating that self-attention is not critical for these datasets. This suggests that convolutional and recurrent layers might already capture the necessary dependencies, reducing the need for self-attention. The removal of the FC layers does not show a clear pattern of degradation. In some cases, there is a tiny improvement in the F1-Score, suggesting that FC layers are not crucial for feature representation and could be simplified. Unlike in DeepConvLSTM, where removing LSTMs had a minimal effect, here, removing the LSTM block results in a moderate performance decrease across most datasets. The aggregation block does not significantly affect performance and the variation is almost negligible across all datasets. Those findings indicate that CNNs are the backbone of TinyHAR, and self-attention and fully connected layers are not essential (in some cases, their removal slightly improves performance); LSTM layers play a moderate role, and aggregation layers contribute almost nothing. In summary, the TinyHAR model relies heavily on convolutional layers, with other components contributing marginally or even negatively in some cases. This suggests that simpler architectures (e.g., CNN-based models without complex attention or aggregation mechanisms) might be sufficient for many HAR tasks, reducing computational overhead without compromising performance.

\subsection{Scalability Exploration}
\label{scalingstudy}

Figure \ref{scaling} illustrates the relationship between model configuration, computational cost, and performance across four datasets: DG, USCHAD, RM, and RECGYM. With respect to the F1-Score, as M and N increase, the F1-Score generally improves across all datasets but with diminishing returns at higher configurations. This trend suggests that larger models tend to perform better, but after a certain point, the performance gains plateau, indicating that further increases in complexity may not yield significant benefits. The result also indicate that the impact of scaling up M and N varies by dataset: for DG and USCHAD, gradual improvement is observed with larger models, but the gains are small beyond a certain point; For RW, the performance fluctuates slightly, suggesting its sensitivity to hyperparameter tuning; For RECGYM, it shows a high F1-Scores even at lower configurations, indicating that a smaller model may be sufficient for this dataset. With respect to MACs, it is clear that MACs increase significantly with larger M and N values, especially as the GRU hidden size (N) increases. With respect to Params, a growing trend is also observed with increasing M and N. 
This scalability exploration indicates that balancing complexity and efficiency is crucial for HAR: while increasing M and N improves performance, it also significantly raises MACs and Params. Larger models perform better in general, but some datasets (e.g., RECGYM) reach near-optimal performance with smaller configurations. Such observations exist commonly in other datasets, suggesting that dataset-dependent tuning is necessary: certain datasets benefit more from increased complexity, while others achieve strong results with smaller architectures. For practical deployment, choosing the smallest configuration that achieves a satisfactory F1-Score can optimize efficiency without excessive computational cost. Based on the global optimized scale across the 14 datasets, we finally adopted 4 and 16 for M and N to present the general optimal choice.

\subsection{Findings and Future Directions}
\label{findings}
\noindent\textbf{Finding 1} Transformer Architecture vs. Classical Architecture. Transformer architectures have gained prominence as powerful tools for time-series classification, primarily due to their ability to capture long-range dependencies and complex temporal patterns. Their attention mechanism enables the dynamic weighting of different input segments, enhancing feature representation. However, our study reveals that self-attention in TinyHAR did not provide significant improvements; in fact, its removal led to better performance. This finding aligns with prior research, such as d’Ascoli et al. from Meta \cite{d2021convit} concluding that convolutional constraints can enable strongly sample-efficient training in the small-data regime than the transformer architecture. A fresh HAR-specific study from Aalto University (Lattanzi et al. (2024) \cite{lattanzi2024transformers} ) also conducted extensive evaluations and observed that transformer-based models consistently yield inferior performance and significantly degrade for edge deployment in the HAR area.
These insights suggest that while transformers offer theoretical advantages, their practical effectiveness in tiny-device applications remains questionable. 

\noindent\textbf{Finding 2} Dataset Diversity of HAR. Across the three models, it has been observed that the same model configuration performs optimally on some datasets while negatively degrading the classification on other datasets. We concluded several key factors that contribute to this inconsistency: (1) The variability in sensor modalities, like the Multi-modal datasets (e.g., OPPO combines motion, audio, and environmental sensors) and wearable sensor datasets (e.g., PAMAP2) primarily use IMUs. A model optimized for multi-modal inputs may struggle with single-sensor datasets due to its reliance on missing features. Conversely, a model fine-tuned for simpler sensor data might not effectively leverage the richness of multi-sensor datasets; (2) The variability in activity type and complexity, like simple activities (e.g., walking, sitting, biking in SHO), which have clear motion patterns, making them easier to classify with CNN-only models, while complex activities (e.g., cooking, cleaning in OPPO) involve multi-step, overlapping motions, requiring advanced models with memory components (RNNs). 
These issues are magnified by annotation ambiguities and label noise \cite{geissler2024beyond}

\noindent\textbf{Finding 3} 
Despite having substantially smaller number of parameters and operations, we have shown that depth-wise separable convolutions can serve as building blocks for HAR, achieving better or comparable performance than their vanilla counterparts. This may be due to their ability to combine temporal and spatial operations, as each filter focuses on a single channel, with later information of all channels being merged by (1,1) convolutions. This also offers more flexibility on when and how to combine the different channels, a process that is typically done through self-attention in TinyHAR or other transformer-based methods such as LimuBert\cite{miao2024spatial}.

\noindent\textbf{Future Direction 1: Pre-training.} Study how pre-training affects different model architectures.  Wearable sensor data has been found to scale \cite{hoddes2025scaling} with self-attention in the presence of enough unlabeled data. This architecture could be compared with other pre-training strategies \cite{hong2024crosshar,thukral2025cross} in the context of resource-constrained encoders. 

\noindent\textbf{Future Direction 2: Dataset-Aware Model Specialization.} Dataset diversity necessitates architectures that adapt to sensor configurations \cite{mhalla2024domain,liu2025assessing}. Thus, for example, modular networks can be designed with composable blocks for auto-configuration based on input streams; And meta-models can be trained on diverse HAR datasets to enable rapid fine-tuning for unseen sensor configurations.



\noindent\textbf{Future Direction 3: Tiered Scaling Framework.} Given the observation that well-designed models can retain competitive accuracy even under substantial reductions in complexity, there is a clear need for a systematic framework (centered on algorithmic scalability) that enables the automated development of ultra-lightweight HAR architectures without compromising performance.

\section{Conclusion}

This work mainly introduces TinierHAR, which redefines the trade-off between efficiency and performance in HAR, achieving unparalleled computational savings while maintaining recognition accuracy. Its lightweight architecture, designed for scalability and adaptability, makes it an optimal solution for real-world deployment on edge devices. The evaluation results confirm that TinierHAR is a superior alternative to conventional HAR models, offering the best balance of efficiency, accuracy, and practical deployability.

\bibliographystyle{ACM-Reference-Format}
\balance
\bibliography{sample-base}


\begin{thebibliography}{52}


\ifx \showCODEN    \undefined \def \showCODEN     #1{\unskip}     \fi
\ifx \showDOI      \undefined \def \showDOI       #1{#1}\fi
\ifx \showISBNx    \undefined \def \showISBNx     #1{\unskip}     \fi
\ifx \showISBNxiii \undefined \def \showISBNxiii  #1{\unskip}     \fi
\ifx \showISSN     \undefined \def \showISSN      #1{\unskip}     \fi
\ifx \showLCCN     \undefined \def \showLCCN      #1{\unskip}     \fi
\ifx \shownote     \undefined \def \shownote      #1{#1}          \fi
\ifx \showarticletitle \undefined \def \showarticletitle #1{#1}   \fi
\ifx \showURL      \undefined \def \showURL       {\relax}        \fi
\providecommand\bibfield[2]{#2}
\providecommand\bibinfo[2]{#2}
\providecommand\natexlab[1]{#1}
\providecommand\showeprint[2][]{arXiv:#2}

\bibitem[DG_({})]%
        {DG_Dataset}
\bibinfo{year}{}\natexlab{}.
\newblock \bibinfo{title}{DG Dataset}.
\newblock \bibinfo{howpublished}{\url{https://archive.ics.uci.edu/dataset/245/daphnet+freezing+of+gait}}.
\newblock
\newblock
\shownote{Accessed: 2025-04-30}.


\bibitem[DSA({})]%
        {DSADS_Dataset}
\bibinfo{year}{}\natexlab{}.
\newblock \bibinfo{title}{DSADS Dataset}.
\newblock \bibinfo{howpublished}{\url{https://archive.ics.uci.edu/dataset/256/daily+and+sports+activities}}.
\newblock
\newblock
\shownote{Accessed: 2025-04-30}.


\bibitem[HAP({})]%
        {HAPT_Dataset}
\bibinfo{year}{}\natexlab{}.
\newblock \bibinfo{title}{HAPT Dataset}.
\newblock \bibinfo{howpublished}{\url{https://archive.ics.uci.edu/dataset/341/smartphone+based+recognition+of+human+activities+and+postural+transitions}}.
\newblock
\newblock
\shownote{Accessed: 2025-04-30}.


\bibitem[MHE({})]%
        {MHEALTH_Dataset}
\bibinfo{year}{}\natexlab{}.
\newblock \bibinfo{title}{MHEALTH Dataset}.
\newblock \bibinfo{howpublished}{\url{https://archive.ics.uci.edu/dataset/319/mhealth+dataset}}.
\newblock
\newblock
\shownote{Accessed: 2025-04-30}.


\bibitem[MOT({})]%
        {MOTIONSENSE_Dataset}
\bibinfo{year}{}\natexlab{}.
\newblock \bibinfo{title}{MOTIONSENSE Dataset}.
\newblock \bibinfo{howpublished}{\url{https://www.kaggle.com/datasets/malekzadeh/motionsense-dataset}}.
\newblock
\newblock
\shownote{Accessed: 2025-04-30}.


\bibitem[OPP({})]%
        {OPPO_Dataset}
\bibinfo{year}{}\natexlab{}.
\newblock \bibinfo{title}{OPPO Dataset}.
\newblock \bibinfo{howpublished}{\url{https://archive.ics.uci.edu/dataset/226/opportunity+activity+recognition}}.
\newblock
\newblock
\shownote{Accessed: 2025-04-30}.


\bibitem[PAM({})]%
        {PAMAP2_Dataset}
\bibinfo{year}{}\natexlab{}.
\newblock \bibinfo{title}{PAMAP2 Dataset}.
\newblock \bibinfo{howpublished}{\url{https://archive.ics.uci.edu/dataset/231/pamap2+physical+activity+monitoring}}.
\newblock
\newblock
\shownote{Accessed: 2025-04-30}.


\bibitem[REC({})]%
        {RECGYM_Dataset}
\bibinfo{year}{}\natexlab{}.
\newblock \bibinfo{title}{RECGYM Dataset}.
\newblock \bibinfo{howpublished}{\url{https://zhaxidele.github.io/RecGym/}}.
\newblock
\newblock
\shownote{Accessed: 2025-04-30}.


\bibitem[RW_({})]%
        {RW_Dataset}
\bibinfo{year}{}\natexlab{}.
\newblock \bibinfo{title}{RW Dataset}.
\newblock \bibinfo{howpublished}{\url{https://www.uni-mannheim.de/dws/research/projects/activity-recognition/dataset/dataset-realworld/}}.
\newblock
\newblock
\shownote{Accessed: 2025-04-30}.


\bibitem[SHO({})]%
        {SHO_Dataset}
\bibinfo{year}{}\natexlab{}.
\newblock \bibinfo{title}{SHO Dataset}.
\newblock \bibinfo{howpublished}{\url{https://www.utwente.nl/en/eemcs/ps/research/dataset/}}.
\newblock
\newblock
\shownote{Accessed: 2025-04-30}.


\bibitem[SKO({})]%
        {SKODAR_Dataset}
\bibinfo{year}{}\natexlab{}.
\newblock \bibinfo{title}{SKODAR Dataset}.
\newblock \bibinfo{howpublished}{\url{http://har-dataset.org/doku.php?id=wiki:dataset}}.
\newblock
\newblock
\shownote{Accessed: 2025-04-30}.


\bibitem[UCI({})]%
        {UCI_Dataset}
\bibinfo{year}{}\natexlab{}.
\newblock \bibinfo{title}{UCI Dataset}.
\newblock \bibinfo{howpublished}{\url{https://archive.ics.uci.edu/dataset/240/human+activity+recognition+using+smartphones}}.
\newblock
\newblock
\shownote{Accessed: 2025-04-30}.


\bibitem[USC({})]%
        {USCHAD_Dataset}
\bibinfo{year}{}\natexlab{}.
\newblock \bibinfo{title}{USCHAD Dataset}.
\newblock \bibinfo{howpublished}{\url{https://sipi.usc.edu/had/}}.
\newblock
\newblock
\shownote{Accessed: 2025-04-30}.


\bibitem[WIS({})]%
        {WISDM_Dataset}
\bibinfo{year}{}\natexlab{}.
\newblock \bibinfo{title}{WISDM Dataset}.
\newblock \bibinfo{howpublished}{\url{https://archive.ics.uci.edu/dataset/507/wisdm+smartphone+and+smartwatch+activity+and+biometrics+dataset}}.
\newblock
\newblock
\shownote{Accessed: 2025-04-30}.


\bibitem[Amft et~al\mbox{.}(2020)]%
        {amft2020wearables}
\bibfield{author}{\bibinfo{person}{Oliver Amft}, \bibinfo{person}{Luis Ignacio~Lopera Gonz{\'a}lez}, \bibinfo{person}{Paul Lukowicz}, \bibinfo{person}{Sizhen Bian}, {and} \bibinfo{person}{Paul Burggraf}.} \bibinfo{year}{2020}\natexlab{}.
\newblock \showarticletitle{Wearables to fight COVID-19: From symptom tracking to contact tracing}.
\newblock \bibinfo{journal}{\emph{IEEE Pervasive Computing}} \bibinfo{volume}{19}, \bibinfo{number}{4} (\bibinfo{year}{2020}), \bibinfo{pages}{53--60}.
\newblock


\bibitem[Batool et~al\mbox{.}(2024)]%
        {batool2024ensemble}
\bibfield{author}{\bibinfo{person}{Sheeza Batool}, \bibinfo{person}{Muhammad~Hassan Khan}, {and} \bibinfo{person}{Muhammad~Shahid Farid}.} \bibinfo{year}{2024}\natexlab{}.
\newblock \showarticletitle{An ensemble deep learning model for human activity analysis using wearable sensory data}.
\newblock \bibinfo{journal}{\emph{Applied Soft Computing}}  \bibinfo{volume}{159} (\bibinfo{year}{2024}), \bibinfo{pages}{111599}.
\newblock


\bibitem[Bian et~al\mbox{.}(2024)]%
        {bian2024device}
\bibfield{author}{\bibinfo{person}{Sizhen Bian}, \bibinfo{person}{Pixi Kang}, \bibinfo{person}{Julian Moosmann}, \bibinfo{person}{Mengxi Liu}, \bibinfo{person}{Pietro Bonazzi}, \bibinfo{person}{Roman Rosipal}, {and} \bibinfo{person}{Michele Magno}.} \bibinfo{year}{2024}\natexlab{}.
\newblock \showarticletitle{On-device Learning of EEGNet-based Network For Wearable Motor Imagery Brain-Computer Interface}. In \bibinfo{booktitle}{\emph{Proceedings of the 2024 ACM International Symposium on Wearable Computers}}. \bibinfo{pages}{9--16}.
\newblock


\bibitem[Bian and Lukowicz(2021a)]%
        {bian2021capacitive}
\bibfield{author}{\bibinfo{person}{Sizhen Bian} {and} \bibinfo{person}{Paul Lukowicz}.} \bibinfo{year}{2021}\natexlab{a}.
\newblock \showarticletitle{Capacitive sensing based on-board hand gesture recognition with tinyml}. In \bibinfo{booktitle}{\emph{Adjunct Proceedings of the 2021 ACM International Joint Conference on Pervasive and Ubiquitous Computing and Proceedings of the 2021 ACM International Symposium on Wearable Computers}}. \bibinfo{pages}{4--5}.
\newblock


\bibitem[Bian and Lukowicz(2021b)]%
        {bian2021systematic}
\bibfield{author}{\bibinfo{person}{Sizhen Bian} {and} \bibinfo{person}{Paul Lukowicz}.} \bibinfo{year}{2021}\natexlab{b}.
\newblock \showarticletitle{A systematic study of the influence of various user specific and environmental factors on wearable human body capacitance sensing}. In \bibinfo{booktitle}{\emph{EAI International Conference on Body Area Networks}}. Springer, \bibinfo{pages}{247--274}.
\newblock


\bibitem[Bian et~al\mbox{.}(2019)]%
        {bian2019passive}
\bibfield{author}{\bibinfo{person}{Sizhen Bian}, \bibinfo{person}{Vitor~F Rey}, \bibinfo{person}{Peter Hevesi}, {and} \bibinfo{person}{Paul Lukowicz}.} \bibinfo{year}{2019}\natexlab{}.
\newblock \showarticletitle{Passive capacitive based approach for full body gym workout recognition and counting}. In \bibinfo{booktitle}{\emph{2019 IEEE International Conference on Pervasive Computing and Communications (PerCom}}. IEEE, \bibinfo{pages}{1--10}.
\newblock


\bibitem[Bian et~al\mbox{.}(2025)]%
        {bian2025hybrid}
\bibfield{author}{\bibinfo{person}{Sizhen Bian}, \bibinfo{person}{Vitor~Fortes Rey}, \bibinfo{person}{Siyu Yuan}, {and} \bibinfo{person}{Paul Lukowicz}.} \bibinfo{year}{2025}\natexlab{}.
\newblock \showarticletitle{Hybrid CNN-Dilated Self-attention Model Using Inertial and Body-Area Electrostatic Sensing for Gym Workout Recognition, Counting, and User Authentification}.
\newblock \bibinfo{journal}{\emph{arXiv preprint arXiv:2503.06311}} (\bibinfo{year}{2025}).
\newblock


\bibitem[Bonazzi et~al\mbox{.}(2024)]%
        {bonazzi2024retina}
\bibfield{author}{\bibinfo{person}{Pietro Bonazzi}, \bibinfo{person}{Sizhen Bian}, \bibinfo{person}{Giovanni Lippolis}, \bibinfo{person}{Yawei Li}, \bibinfo{person}{Sadique Sheik}, {and} \bibinfo{person}{Michele Magno}.} \bibinfo{year}{2024}\natexlab{}.
\newblock \showarticletitle{Retina: Low-power eye tracking with event camera and spiking hardware}. In \bibinfo{booktitle}{\emph{Proceedings of the IEEE/CVF Conference on Computer Vision and Pattern Recognition}}. \bibinfo{pages}{5684--5692}.
\newblock


\bibitem[Deng et~al\mbox{.}(2023)]%
        {deng2023lhar}
\bibfield{author}{\bibinfo{person}{Shizhuo Deng}, \bibinfo{person}{Jiaqi Chen}, \bibinfo{person}{Da Teng}, \bibinfo{person}{Chuangui Yang}, \bibinfo{person}{Dongyue Chen}, \bibinfo{person}{Tong Jia}, {and} \bibinfo{person}{Hao Wang}.} \bibinfo{year}{2023}\natexlab{}.
\newblock \showarticletitle{Lhar: Lightweight human activity recognition on knowledge distillation}.
\newblock \bibinfo{journal}{\emph{IEEE Journal of Biomedical and Health Informatics}} (\bibinfo{year}{2023}).
\newblock


\bibitem[d’Ascoli et~al\mbox{.}(2021)]%
        {d2021convit}
\bibfield{author}{\bibinfo{person}{St{\'e}phane d’Ascoli}, \bibinfo{person}{Hugo Touvron}, \bibinfo{person}{Matthew~L Leavitt}, \bibinfo{person}{Ari~S Morcos}, \bibinfo{person}{Giulio Biroli}, {and} \bibinfo{person}{Levent Sagun}.} \bibinfo{year}{2021}\natexlab{}.
\newblock \showarticletitle{Convit: Improving vision transformers with soft convolutional inductive biases}. In \bibinfo{booktitle}{\emph{International conference on machine learning}}. PMLR, \bibinfo{pages}{2286--2296}.
\newblock


\bibitem[Fiorini et~al\mbox{.}(2021)]%
        {fiorini2021daily}
\bibfield{author}{\bibinfo{person}{Laura Fiorini}, \bibinfo{person}{Federica G~Cornacchia Loizzo}, \bibinfo{person}{Alessandra Sorrentino}, \bibinfo{person}{Jaeseok Kim}, \bibinfo{person}{Erika Rovini}, \bibinfo{person}{Alessandro Di~Nuovo}, {and} \bibinfo{person}{Filippo Cavallo}.} \bibinfo{year}{2021}\natexlab{}.
\newblock \showarticletitle{Daily gesture recognition during human-robot interaction combining vision and wearable systems}.
\newblock \bibinfo{journal}{\emph{IEEE Sensors Journal}} \bibinfo{volume}{21}, \bibinfo{number}{20} (\bibinfo{year}{2021}), \bibinfo{pages}{23568--23577}.
\newblock


\bibitem[Geissler et~al\mbox{.}(2024)]%
        {geissler2024beyond}
\bibfield{author}{\bibinfo{person}{Daniel Geissler}, \bibinfo{person}{Dominique Nshimyimana}, \bibinfo{person}{Vitor~Fortes Rey}, \bibinfo{person}{Sungho Suh}, \bibinfo{person}{Bo Zhou}, {and} \bibinfo{person}{Paul Lukowicz}.} \bibinfo{year}{2024}\natexlab{}.
\newblock \showarticletitle{Beyond Confusion: A Fine-grained Dialectical Examination of Human Activity Recognition Benchmark Datasets}.
\newblock \bibinfo{journal}{\emph{arXiv preprint arXiv:2412.09037}} (\bibinfo{year}{2024}).
\newblock


\bibitem[Hamad et~al\mbox{.}(2021)]%
        {hamad2021dilated}
\bibfield{author}{\bibinfo{person}{Rebeen~Ali Hamad}, \bibinfo{person}{Masashi Kimura}, \bibinfo{person}{Longzhi Yang}, \bibinfo{person}{Wai~Lok Woo}, {and} \bibinfo{person}{Bo Wei}.} \bibinfo{year}{2021}\natexlab{}.
\newblock \showarticletitle{Dilated causal convolution with multi-head self attention for sensor human activity recognition}.
\newblock \bibinfo{journal}{\emph{Neural Computing and Applications}}  \bibinfo{volume}{33} (\bibinfo{year}{2021}), \bibinfo{pages}{13705--13722}.
\newblock


\bibitem[Hoddes et~al\mbox{.}(2025)]%
        {hoddes2025scaling}
\bibfield{author}{\bibinfo{person}{Tom Hoddes}, \bibinfo{person}{Alex Bijamov}, \bibinfo{person}{Saket Joshi}, \bibinfo{person}{Daniel Roggen}, \bibinfo{person}{Ali Etemad}, \bibinfo{person}{Robert Harle}, {and} \bibinfo{person}{David Racz}.} \bibinfo{year}{2025}\natexlab{}.
\newblock \showarticletitle{Scaling laws in wearable human activity recognition}.
\newblock \bibinfo{journal}{\emph{arXiv preprint arXiv:2502.03364}} (\bibinfo{year}{2025}).
\newblock


\bibitem[Hong et~al\mbox{.}(2024)]%
        {hong2024crosshar}
\bibfield{author}{\bibinfo{person}{Zhiqing Hong}, \bibinfo{person}{Zelong Li}, \bibinfo{person}{Shuxin Zhong}, \bibinfo{person}{Wenjun Lyu}, \bibinfo{person}{Haotian Wang}, \bibinfo{person}{Yi Ding}, \bibinfo{person}{Tian He}, {and} \bibinfo{person}{Desheng Zhang}.} \bibinfo{year}{2024}\natexlab{}.
\newblock \showarticletitle{Crosshar: Generalizing cross-dataset human activity recognition via hierarchical self-supervised pretraining}.
\newblock \bibinfo{journal}{\emph{Proceedings of the ACM on Interactive, Mobile, Wearable and Ubiquitous Technologies}} \bibinfo{volume}{8}, \bibinfo{number}{2} (\bibinfo{year}{2024}), \bibinfo{pages}{1--26}.
\newblock


\bibitem[Igwe et~al\mbox{.}(2022)]%
        {igwe2022human}
\bibfield{author}{\bibinfo{person}{Ogbonna~Michael Igwe}, \bibinfo{person}{Yi Wang}, \bibinfo{person}{George~C Giakos}, {and} \bibinfo{person}{Jian Fu}.} \bibinfo{year}{2022}\natexlab{}.
\newblock \showarticletitle{Human activity recognition in smart environments employing margin setting algorithm}.
\newblock \bibinfo{journal}{\emph{Journal of Ambient Intelligence and Humanized Computing}} \bibinfo{volume}{13}, \bibinfo{number}{7} (\bibinfo{year}{2022}), \bibinfo{pages}{3669--3681}.
\newblock


\bibitem[Kang et~al\mbox{.}(2025)]%
        {kang2025human}
\bibfield{author}{\bibinfo{person}{Hari Kang}, \bibinfo{person}{Donghyun Kim}, {and} \bibinfo{person}{Kar-Ann Toh}.} \bibinfo{year}{2025}\natexlab{}.
\newblock \showarticletitle{Human Activity Recognition Through Augmented WiFi CSI Signals by Lightweight Attention-GRU}.
\newblock \bibinfo{journal}{\emph{Sensors}} \bibinfo{volume}{25}, \bibinfo{number}{5} (\bibinfo{year}{2025}), \bibinfo{pages}{1547}.
\newblock


\bibitem[Kang et~al\mbox{.}(2024)]%
        {kang2024device}
\bibfield{author}{\bibinfo{person}{Pixi Kang}, \bibinfo{person}{Julian Moosmann}, \bibinfo{person}{Sizhen Bian}, {and} \bibinfo{person}{Michele Magno}.} \bibinfo{year}{2024}\natexlab{}.
\newblock \showarticletitle{On-Device Training Empowered Transfer Learning For Human Activity Recognition}.
\newblock \bibinfo{journal}{\emph{arXiv preprint arXiv:2407.03644}} (\bibinfo{year}{2024}).
\newblock


\bibitem[Lalwani and Ganeshan(2024)]%
        {lalwani2024novel}
\bibfield{author}{\bibinfo{person}{Pooja Lalwani} {and} \bibinfo{person}{R Ganeshan}.} \bibinfo{year}{2024}\natexlab{}.
\newblock \showarticletitle{A Novel CNN-BiLSTM-GRU Hybrid Deep Learning Model for Human Activity Recognition}.
\newblock \bibinfo{journal}{\emph{International Journal of Computational Intelligence Systems}} \bibinfo{volume}{17}, \bibinfo{number}{1} (\bibinfo{year}{2024}), \bibinfo{pages}{1--20}.
\newblock


\bibitem[Lattanzi et~al\mbox{.}(2024)]%
        {lattanzi2024transformers}
\bibfield{author}{\bibinfo{person}{Emanuele Lattanzi}, \bibinfo{person}{Lorenzo Calisti}, {and} \bibinfo{person}{Chiara Contoli}.} \bibinfo{year}{2024}\natexlab{}.
\newblock \showarticletitle{Are Transformers a Useful Tool for Tiny devices in Human Activity Recognition?}. In \bibinfo{booktitle}{\emph{Proceedings of the 2024 8th International Conference on Advances in Artificial Intelligence}}. \bibinfo{pages}{339--344}.
\newblock


\bibitem[Lim et~al\mbox{.}(2023)]%
        {lim2023efficient}
\bibfield{author}{\bibinfo{person}{Won-Seon Lim}, \bibinfo{person}{Wangduk Seo}, \bibinfo{person}{Dae-Won Kim}, {and} \bibinfo{person}{Jaesung Lee}.} \bibinfo{year}{2023}\natexlab{}.
\newblock \showarticletitle{Efficient Human Activity Recognition Using Lookup Table-Based Neural Architecture Search for Mobile Devices}.
\newblock \bibinfo{journal}{\emph{IEEE Access}}  \bibinfo{volume}{11} (\bibinfo{year}{2023}), \bibinfo{pages}{71727--71738}.
\newblock


\bibitem[Liu et~al\mbox{.}(2022)]%
        {liu2022non}
\bibfield{author}{\bibinfo{person}{Mengxi Liu}, \bibinfo{person}{Sizhen Bian}, {and} \bibinfo{person}{Paul Lukowicz}.} \bibinfo{year}{2022}\natexlab{}.
\newblock \showarticletitle{Non-contact, real-time eye blink detection with capacitive sensing}. In \bibinfo{booktitle}{\emph{Proceedings of the 2022 ACM International Symposium on Wearable Computers}}. \bibinfo{pages}{49--53}.
\newblock


\bibitem[Liu et~al\mbox{.}(2025)]%
        {liu2025assessing}
\bibfield{author}{\bibinfo{person}{Mengxi Liu}, \bibinfo{person}{Daniel Gei{\ss}ler}, \bibinfo{person}{Sizhen Bian}, \bibinfo{person}{Bo Zhou}, {and} \bibinfo{person}{Paul Lukowicz}.} \bibinfo{year}{2025}\natexlab{}.
\newblock \showarticletitle{Assessing the Impact of Sampling Irregularity in Time Series Data: Human Activity Recognition As A Case Study}.
\newblock \bibinfo{journal}{\emph{arXiv preprint arXiv:2501.15330}} (\bibinfo{year}{2025}).
\newblock


\bibitem[Luder et~al\mbox{.}(2024)]%
        {luder2024cyclowatt}
\bibfield{author}{\bibinfo{person}{Victor Luder}, \bibinfo{person}{Sizhen Bian}, {and} \bibinfo{person}{Michele Magno}.} \bibinfo{year}{2024}\natexlab{}.
\newblock \showarticletitle{CycloWatt: An Affordable, TinyML-enhanced IoT Device Revolutionizing Cycling Power Metrics}. In \bibinfo{booktitle}{\emph{2024 IEEE International Instrumentation and Measurement Technology Conference (I2MTC)}}. IEEE, \bibinfo{pages}{1--6}.
\newblock


\bibitem[Mhalla and Favreau(2024)]%
        {mhalla2024domain}
\bibfield{author}{\bibinfo{person}{Ala Mhalla} {and} \bibinfo{person}{Jean-Marie Favreau}.} \bibinfo{year}{2024}\natexlab{}.
\newblock \showarticletitle{Domain adaptation framework for personalized human activity recognition models}.
\newblock \bibinfo{journal}{\emph{Multimedia Tools and Applications}} \bibinfo{volume}{83}, \bibinfo{number}{25} (\bibinfo{year}{2024}), \bibinfo{pages}{66775--66797}.
\newblock


\bibitem[Miao et~al\mbox{.}(2024)]%
        {miao2024spatial}
\bibfield{author}{\bibinfo{person}{Shenghuan Miao}, \bibinfo{person}{Ling Chen}, {and} \bibinfo{person}{Rong Hu}.} \bibinfo{year}{2024}\natexlab{}.
\newblock \showarticletitle{Spatial-temporal masked autoencoder for multi-device wearable human activity recognition}.
\newblock \bibinfo{journal}{\emph{Proceedings of the ACM on Interactive, Mobile, Wearable and Ubiquitous Technologies}} \bibinfo{volume}{7}, \bibinfo{number}{4} (\bibinfo{year}{2024}), \bibinfo{pages}{1--25}.
\newblock


\bibitem[Muhoza et~al\mbox{.}(2023)]%
        {muhoza2023power}
\bibfield{author}{\bibinfo{person}{Aim{\'e}~Cedric Muhoza}, \bibinfo{person}{Emmanuel Bergeret}, \bibinfo{person}{Corinne Brdys}, {and} \bibinfo{person}{Francis Gary}.} \bibinfo{year}{2023}\natexlab{}.
\newblock \showarticletitle{Power consumption reduction for IoT devices thanks to Edge-AI: Application to human activity recognition}.
\newblock \bibinfo{journal}{\emph{Internet of Things}}  \bibinfo{volume}{24} (\bibinfo{year}{2023}), \bibinfo{pages}{100930}.
\newblock


\bibitem[Phukan et~al\mbox{.}(2022)]%
        {phukan2022convolutional}
\bibfield{author}{\bibinfo{person}{Nabasmita Phukan}, \bibinfo{person}{Shailesh Mohine}, \bibinfo{person}{Achinta Mondal}, \bibinfo{person}{M~Sabarimalai Manikandan}, {and} \bibinfo{person}{Ram~Bilas Pachori}.} \bibinfo{year}{2022}\natexlab{}.
\newblock \showarticletitle{Convolutional neural network-based human activity recognition for edge fitness and context-aware health monitoring devices}.
\newblock \bibinfo{journal}{\emph{IEEE Sensors Journal}} \bibinfo{volume}{22}, \bibinfo{number}{22} (\bibinfo{year}{2022}), \bibinfo{pages}{21816--21826}.
\newblock


\bibitem[Ronald et~al\mbox{.}(2021)]%
        {ronald2021isplinception}
\bibfield{author}{\bibinfo{person}{Mutegeki Ronald}, \bibinfo{person}{Alwin Poulose}, {and} \bibinfo{person}{Dong~Seog Han}.} \bibinfo{year}{2021}\natexlab{}.
\newblock \showarticletitle{iSPLInception: an inception-ResNet deep learning architecture for human activity recognition}.
\newblock \bibinfo{journal}{\emph{IEEE Access}}  \bibinfo{volume}{9} (\bibinfo{year}{2021}), \bibinfo{pages}{68985--69001}.
\newblock


\bibitem[Serpush et~al\mbox{.}(2022)]%
        {serpush2022wearable}
\bibfield{author}{\bibinfo{person}{Fatemeh Serpush}, \bibinfo{person}{Mohammad~Bagher Menhaj}, \bibinfo{person}{Behrooz Masoumi}, {and} \bibinfo{person}{Babak Karasfi}.} \bibinfo{year}{2022}\natexlab{}.
\newblock \showarticletitle{Wearable sensor-based human activity recognition in the smart healthcare system}.
\newblock \bibinfo{journal}{\emph{Computational intelligence and neuroscience}} \bibinfo{volume}{2022}, \bibinfo{number}{1} (\bibinfo{year}{2022}), \bibinfo{pages}{1391906}.
\newblock


\bibitem[Song et~al\mbox{.}(2018)]%
        {song2018deep}
\bibfield{author}{\bibinfo{person}{Sibo Song}, \bibinfo{person}{Ngai-Man Cheung}, \bibinfo{person}{Vijay Chandrasekhar}, {and} \bibinfo{person}{Bappaditya Mandal}.} \bibinfo{year}{2018}\natexlab{}.
\newblock \showarticletitle{Deep adaptive temporal pooling for activity recognition}. In \bibinfo{booktitle}{\emph{Proceedings of the 26th ACM international conference on Multimedia}}. \bibinfo{pages}{1829--1837}.
\newblock


\bibitem[Thukral et~al\mbox{.}(2025)]%
        {thukral2025cross}
\bibfield{author}{\bibinfo{person}{Megha Thukral}, \bibinfo{person}{Harish Haresamudram}, {and} \bibinfo{person}{Thomas Ploetz}.} \bibinfo{year}{2025}\natexlab{}.
\newblock \showarticletitle{Cross-Domain HAR: Few-Shot Transfer Learning for Human Activity Recognition}.
\newblock \bibinfo{journal}{\emph{ACM Transactions on Intelligent Systems and Technology}} \bibinfo{volume}{16}, \bibinfo{number}{1} (\bibinfo{year}{2025}), \bibinfo{pages}{1--35}.
\newblock


\bibitem[Xu and Qiu(2021)]%
        {xu2021human}
\bibfield{author}{\bibinfo{person}{Yang Xu} {and} \bibinfo{person}{Ting~Ting Qiu}.} \bibinfo{year}{2021}\natexlab{}.
\newblock \showarticletitle{Human activity recognition and embedded application based on convolutional neural network}.
\newblock \bibinfo{journal}{\emph{Journal of Artificial Intelligence and Technology}} \bibinfo{volume}{1}, \bibinfo{number}{1} (\bibinfo{year}{2021}), \bibinfo{pages}{51--60}.
\newblock


\bibitem[Yi et~al\mbox{.}(2023)]%
        {yi2023human}
\bibfield{author}{\bibinfo{person}{Myung-Kyu Yi}, \bibinfo{person}{Wai-Kong Lee}, {and} \bibinfo{person}{Seong~Oun Hwang}.} \bibinfo{year}{2023}\natexlab{}.
\newblock \showarticletitle{A human activity recognition method based on lightweight feature extraction combined with pruned and quantized CNN for wearable device}.
\newblock \bibinfo{journal}{\emph{IEEE Transactions on Consumer Electronics}} \bibinfo{volume}{69}, \bibinfo{number}{3} (\bibinfo{year}{2023}), \bibinfo{pages}{657--670}.
\newblock


\bibitem[Zhang et~al\mbox{.}(2022)]%
        {zhang2022deep}
\bibfield{author}{\bibinfo{person}{Shibo Zhang}, \bibinfo{person}{Yaxuan Li}, \bibinfo{person}{Shen Zhang}, \bibinfo{person}{Farzad Shahabi}, \bibinfo{person}{Stephen Xia}, \bibinfo{person}{Yu Deng}, {and} \bibinfo{person}{Nabil Alshurafa}.} \bibinfo{year}{2022}\natexlab{}.
\newblock \showarticletitle{Deep learning in human activity recognition with wearable sensors: A review on advances}.
\newblock \bibinfo{journal}{\emph{Sensors}} \bibinfo{volume}{22}, \bibinfo{number}{4} (\bibinfo{year}{2022}), \bibinfo{pages}{1476}.
\newblock


\bibitem[Zhou et~al\mbox{.}(2024)]%
        {zhou2024mlp}
\bibfield{author}{\bibinfo{person}{Yexu Zhou}, \bibinfo{person}{Tobias King}, \bibinfo{person}{Haibin Zhao}, \bibinfo{person}{Yiran Huang}, \bibinfo{person}{Till Riedel}, {and} \bibinfo{person}{Michael Beigl}.} \bibinfo{year}{2024}\natexlab{}.
\newblock \showarticletitle{MLP-HAR: Boosting Performance and Efficiency of HAR Models on Edge Devices with Purely Fully Connected Layers}. In \bibinfo{booktitle}{\emph{Proceedings of the 2024 ACM International Symposium on Wearable Computers}}. \bibinfo{pages}{133--139}.
\newblock


\bibitem[Zhou et~al\mbox{.}(2022)]%
        {zhou2022tinyhar}
\bibfield{author}{\bibinfo{person}{Yexu Zhou}, \bibinfo{person}{Haibin Zhao}, \bibinfo{person}{Yiran Huang}, \bibinfo{person}{Till Riedel}, \bibinfo{person}{Michael Hefenbrock}, {and} \bibinfo{person}{Michael Beigl}.} \bibinfo{year}{2022}\natexlab{}.
\newblock \showarticletitle{Tinyhar: A lightweight deep learning model designed for human activity recognition}. In \bibinfo{booktitle}{\emph{Proceedings of the 2022 ACM International Symposium on Wearable Computers}}. \bibinfo{pages}{89--93}.
\newblock


\bibitem[Zhu et~al\mbox{.}(2021)]%
        {zhu2021lightweight}
\bibfield{author}{\bibinfo{person}{Jianping Zhu}, \bibinfo{person}{Xin Lou}, {and} \bibinfo{person}{Wenbin Ye}.} \bibinfo{year}{2021}\natexlab{}.
\newblock \showarticletitle{Lightweight deep learning model in mobile-edge computing for radar-based human activity recognition}.
\newblock \bibinfo{journal}{\emph{IEEE Internet of Things Journal}} \bibinfo{volume}{8}, \bibinfo{number}{15} (\bibinfo{year}{2021}), \bibinfo{pages}{12350--12359}.
\newblock


\end{thebibliography}


\end{document}